\documentclass[sigconf]{acmart}
\AtBeginDocument{%
  }

\setcopyright{acmlicensed}
\copyrightyear{2018}
\acmYear{2018}
\acmDOI{XXXXXXX.XXXXXXX}
\acmConference[Conference acronym 'XX]{Make sure to enter the correct
  conference title from your rights confirmation email}{June 03--05,
  2018}{Woodstock, NY}
\acmISBN{978-1-4503-XXXX-X/2018/06}

\usepackage{booktabs}
\usepackage{array}
\usepackage{multirow}
\usepackage{enumitem}
\usepackage{amsmath}
\usepackage{graphicx}
\usepackage{subcaption}
\usepackage{algorithm}
\usepackage{algpseudocode}

\begin{document}

\title{KARMA: Training-Free Knowledge-Atomic Harness Evolution for Industrial Advertising Agents}


\author{Xinhua Miao}
\authornote{Both authors contributed equally to this research.}
\email{miaoxinhua.mxh@alibaba-inc.com}
\orcid{0009-0002-3620-4147}
\author{Yuanzhen Rong}
\authornotemark[1]
\email{yuanzhen.ryz@alibaba-inc.com}
\orcid{0009-0006-8503-9378}
\affiliation{%
  \institution{Alibaba International Digital Commerce Group}
  \city{Hangzhou}
  \country{China}
}

\author{Weiru Zhang}
\orcid{0009-0007-8496-3638}
\email{weiru.zwr@alibaba-inc.com}
\affiliation{%
  \institution{Alibaba International Digital Commerce Group}
  \city{Hangzhou}
  \country{China}
}

\author{Xiaoyi Zeng}
\orcid{0000-0002-3742-4910}
\authornote{Corresponding author.}
\email{yuanhan@taobao.com}
\affiliation{%
  \institution{Alibaba International Digital Commerce Group}
  \city{Hangzhou}
  \country{China}
}

\renewcommand{\shortauthors}{Miao and Rong et al.}

\begin{abstract}
Industrial advertising agents must continually adapt their decision knowledge to changing product offerings, traffic patterns, and operating policies. We present \textbf{KARMA} (\textbf{K}nowledge-\textbf{A}tomic, \textbf{R}egression-Gated \textbf{M}odular \textbf{A}daptation), a training-free framework that evolves six typed knowledge modules while keeping model weights and executable harness code frozen. During our evolution, a deterministic Trace Verifier maps observed errors to candidate modules, and a Critic compares successful and failed trajectories to guide a tool-constrained Evolver. The Evolver applies atomic knowledge edits, which a fixed validation gate retains only when their measured gain reaches a threshold; otherwise, the previous knowledge state is restored. We evaluate KARMA with three LLM families on cases sampled from a corpus of approximately half a million real-world operation logs across six e-commerce domains. Across 18 domain--model settings, KARMA raises the average hold-out score from 30.19 to 39.15 (+8.96 points, +29.7\%). Ablations show that removing either successful or failed trajectories more than halves the average hold-out gain. Evolved knowledge also transfers across domains without further adaptation, yielding an average hold-out gain of 8.54 points across all source--target pairs. By linking retained patches to their diagnostic evidence and validation outcomes, KARMA supports inspectable, reversible knowledge adaptation for agents in Web marketplaces.
\end{abstract}




\maketitle

\section{Introduction}
Web marketplaces allocate merchant spending and product exposure through advertising systems. Merchants must decide which products to promote, how much budget to allocate, and how aggressively to bid under changing demand and competition. LLM agents can retrieve business evidence and combine it with operating knowledge~\cite{rtbagent}, but their recommendations depend on rules that may become incomplete or outdated. Updating those rules can change both the selected control (budget, bid, or item) and the direction of intervention. The research challenge is to improve this decision knowledge while making changes inspectable and screening for regressions across marketplace contexts. We study this challenge through an advertising agent that queries telemetry from a global e-commerce platform and emits structured decisions in an offline evaluation.

\begin{figure*}[t]
\centering
\includegraphics[width=0.98\linewidth]{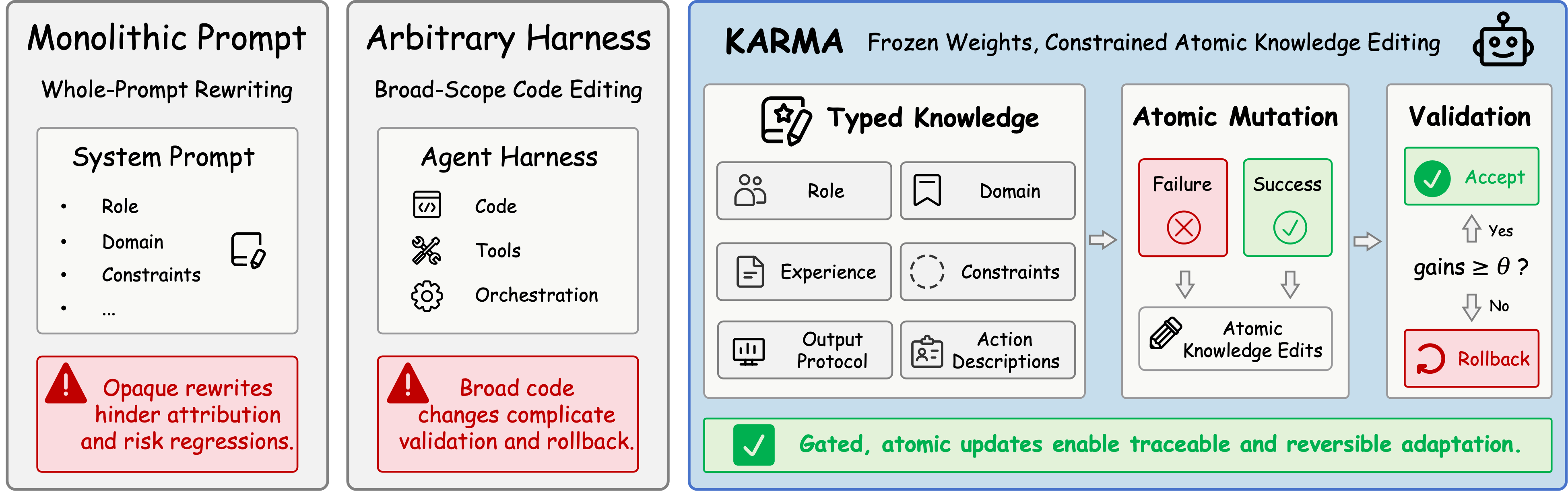}
\caption{Motivation. Whole-prompt rewriting obscures individual changes; broad harness editing complicates validation and rollback. KARMA freezes weights and code, derives atomic knowledge edits from successes and failures, and retains only patches whose validation gain reaches $\theta$.}
\Description{Three panels. The left panel shows a monolithic system prompt containing role, domain, and constraints, with the warning that opaque rewrites hinder attribution and risk regressions. The middle panel shows an agent harness containing code, tools, and orchestration, with the warning that broad code changes complicate validation and rollback. The right panel shows KARMA: six typed knowledge modules, atomic knowledge edits derived from failures and successes, and a validation step that accepts an update if gains are at least theta and otherwise rolls it back.}
\label{fig:motivation}
\end{figure*}

Repeated fine-tuning is costly, while manual prompt maintenance is difficult to scale as products and policies change. Whole-prompt rewriting can entangle changes to domain knowledge, decision heuristics, and output instructions. Broad harness-code editing provides more flexibility but also changes the tools or control flow that must be validated (Figure~\ref{fig:motivation}). This motivates a narrower adaptation object: external knowledge that can be revised while the executable decision process remains fixed. We therefore ask: \emph{can an industrial advertising agent improve its decision knowledge from evaluated trajectories without changing model weights or executable harness code?}

Existing self-improving agents revise prompts, memories, workflows, or harnesses, mostly in coding and related environments~\cite{gepa,ahe,selfharness,metaharness,harnessfix,harnesscompass,starharness}. Advertising introduces three considerations for this process. First, a wrong action type differs from a wrong direction within the correct type, so feedback should preserve this structure. Second, interacting rules can overgeneralize: a repair that helps failed cases may disrupt correct decisions. Third, tool and protocol failures must be distinguished from semantic errors before proposing knowledge changes. These considerations motivate structured diagnosis, evidence from both successes and failures, and an outcome-based acceptance gate.

We introduce \textbf{KARMA} (\textbf{K}nowledge-\textbf{A}tomic, \textbf{R}egression-Gated \textbf{M}odular \textbf{A}daptation), a training-free framework that externalizes knowledge into six typed modules and restricts adaptation to schema-checked, journaled mutations. Rule modules preserve stable identifiers and provenance, allowing an edit to be inspected alongside the evidence that motivated it. The full knowledge context remains available during inference, while model weights, business tools, and executable code stay fixed. Each edit has an explicit target, and validation measures its effect on agent decisions.

A deterministic Trace Verifier maps observed failures to candidate knowledge modules and retains successful trajectories as context for repair. A Critic compares these evidence channels, and a tool-constrained Evolver translates suitable proposals into a small patch. The outer loop retains a patch only if its gain on a fixed validation set reaches $\theta$, and otherwise restores the full knowledge snapshot. An untouched hold-out set measures final generalization. Each cycle records the diagnostic evidence, knowledge changes, and validation outcome.

Our contributions are summarized as follows:
\begin{itemize}
  \item To our knowledge, we present the first training-free harness self-evolution framework designed and evaluated for an industrial advertising decision agent.
  \item We make typed, rule-addressable knowledge the object of evolution. Schema-checked atomic edits, provenance, mutation journals, and full-directory snapshots support inspection and rollback while preserving the executable harness.
  \item We combine deterministic component attribution with full success/failure ingestion: failures guide repair, while successes provide evidence for preserving effective behavior.
  \item Across six domains and three LLM families, KARMA improves the average hold-out score by 8.96 points (+29.7\%). Ablations, engine comparisons, and cross-domain transfers examine the design and its generalization.
\end{itemize}

\begin{figure*}[t]
\centering
\includegraphics[width=0.98\linewidth]{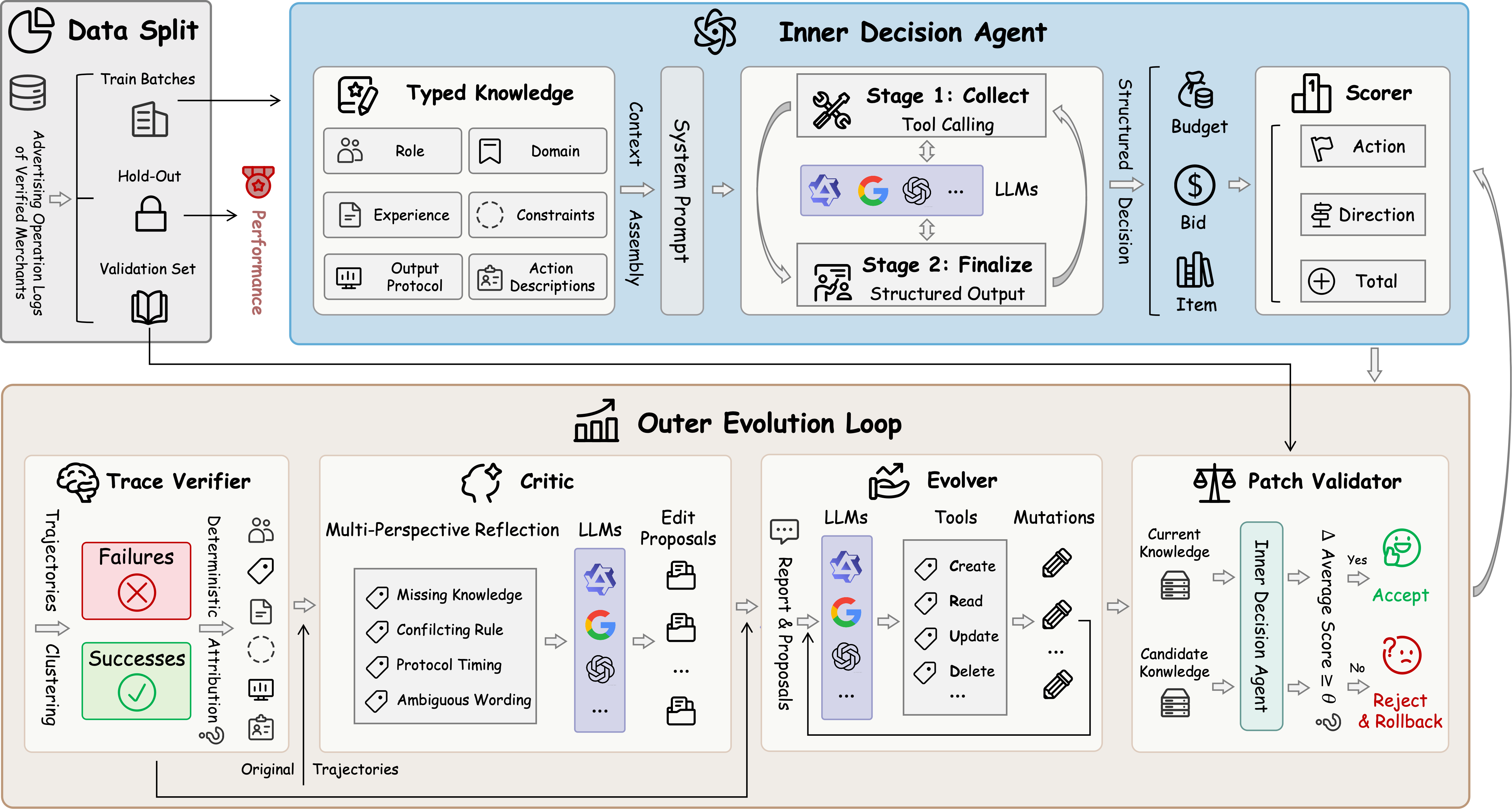}
\caption{KARMA overview. The frozen inner agent collects evidence and emits structured decisions. The outer loop converts verified trajectories into atomic knowledge edits through the Critic and Evolver, then validates or rolls back each patch. The hold-out set is excluded from evolution.}
\Description{The top half shows advertising operation logs of verified merchants split into training batches, a hold-out set used for final performance, and a validation set. The inner decision agent renders six typed knowledge modules into the context and response schema, runs a two-stage loop of tool calling and structured output with LLMs, emits budget, bid, or item decisions, and scores them by action, direction, and total. The bottom half shows the outer evolution loop: a Trace Verifier clusters failures and successes with deterministic attribution; a Critic performs multi-perspective reflection over missing knowledge, conflicting rules, protocol timing, and ambiguous wording to produce edit proposals; an Evolver uses create, read, update, and delete tools to produce mutations; and a Patch Validator compares current and candidate knowledge on the validation set, accepting if the average score gain is at least theta and otherwise rejecting and rolling back.}
\label{fig:overview}
\end{figure*}

\section{Related Work}
\subsection{Training-Free Agent Adaptation}
Tool-using agents combine language-model reasoning with external observations~\cite{react,toolformer}. Reflexion and Self-Refine use verbal feedback to improve subsequent attempts or iteratively revise outputs without additional weight updates~\cite{reflexion,selfrefine}. Prompt optimizers update instructions using task feedback: ProTeGi combines textual critiques with beam search~\cite{protegi}, OPRO generates candidates from previously evaluated solutions~\cite{opro}, and GEPA combines reflective prompt updates through Pareto selection~\cite{gepa}.

Persistent memories and skills offer another adaptation mechanism. ExpeL extracts reusable insights from successful and failed experiences~\cite{expel}, while Voyager accumulates executable skills through environment feedback and self-verification~\cite{voyager}. ACE maintains structured context playbooks through generation, reflection, and curation, using incremental updates to preserve accumulated knowledge~\cite{ace}. RHO revisits challenging past tasks and selects harness updates through pairwise self-preference without labeled validation data~\cite{rho}. RSIAgent constructs environment-specific memory through curriculum, actor, and verifier roles~\cite{rsiagent}, while MetaSkill-Evolve adapts both task skills and the procedure that evolves them~\cite{metaskill}.

However, linking feedback to specific knowledge components and preserving effective behavior during revision remain challenging when decision rules interact.

\subsection{Harness Evolution and Repair}
Beyond individual prompts, DSPy optimizes modular language-model pipelines against task metrics~\cite{dspy}, and TextGrad propagates textual feedback through computation graphs~\cite{textgrad}. ADAS uses a meta agent to search code-defined agent designs~\cite{adas}; AFlow searches executable workflows using Monte Carlo Tree Search and execution feedback~\cite{aflow}.

Harness evolution extends adaptation to the runtime surrounding a frozen model. AHE exposes editable components, organizes execution evidence, and checks predicted edit outcomes in subsequent iterations~\cite{ahe}. Self-Harness mines failure patterns, proposes minimal modifications, and accepts them after regression testing~\cite{selfharness}. Meta-Harness searches executable harness programs using the code, scores, and traces of prior candidates~\cite{metaharness}; RHI instead refines prompt-level specifications of the agent loop through pairwise feedback over revision history~\cite{rhi}.

Several methods explicitly address diagnosis and generalization. HarnessCompass combines constrained edits, proactive agent feedback, and component-wise optimization~\cite{harnesscompass}. StarHarness stratifies evolution tasks and separates proposer-visible search from hidden selection tasks~\cite{starharness}. HarnessFix aligns traces with implementation artifacts to guide scoped repairs and regression-aware validation~\cite{harnessfix}. HarnessBank maintains diverse candidate harnesses in a gene bank and screens offspring through gated verification~\cite{harnessbank}. Related analyses distinguish the ability to generate useful harness updates from the ability to benefit from them during execution~\cite{harnessbenefit}.

Editing executable harnesses expands the search space and the dependencies that must be validated. Even accepted changes may generalize unevenly across tasks, making evaluation beyond the optimization setting essential.

\subsection{Commerce and Advertising Agents}
Commerce agents span product search, customer interaction, and advertising decisions. WebShop evaluates grounded shopping agents that navigate product pages to satisfy user instructions~\cite{webshop}. In advertising, RTBAgent combines click-through-rate estimates, expert strategies, memory retrieval, and daily reflection to inform bidding decisions~\cite{rtbagent}. HAT addresses changing harness configurations for live-commerce digital avatars through harness-state augmentation, supervised fine-tuning, distillation, and reinforcement learning~\cite{hat}.

These studies focus on specific interaction or bidding tasks; evidence on reusable knowledge updates spanning budget, bid, and item decisions across product domains remains limited.

\section{Problem Formulation}
Let $x$ denote the decision context of an advertising case, containing merchant and campaign identifiers, campaign type, and timestamp. The logged reference action $y^*$ and its details are kept separate from $x$ and are excluded from the inner agent's inputs and tool observations. Reference actions are used only for scoring and outer-loop feedback after inference. The inner agent $A_{\phi}(x;K)$ has frozen model parameters $\phi$ and a mutable external knowledge state $K$. It queries read-only business tools and emits a structured decision
\begin{equation}
 y=(a,d,r,E), \quad a \in \{\texttt{budget},\texttt{bid},\texttt{item}\},
\end{equation}
where $a$ is the action type, $d$ the action direction, $r$ a rationale, and $E$ contains evidence identifiers. Budget and bid actions select \texttt{increase} or \texttt{decrease}; item actions select add or stop and additionally return item identifiers.

For a logged ground-truth action $y^*$, let $t(\cdot)$ and $d(\cdot)$ denote action type and direction. We use a strict score
\begin{equation}
 s(y,y^*)=\tfrac{1}{2}\mathbb{1}[t(y)=t(y^*)]
 +\tfrac{1}{2}\mathbb{1}[t(y)=t(y^*)\land d(y)=d(y^*)].
 \label{eq:score}
\end{equation}
Here, $\mathbb{1}[\cdot]$ equals one when its condition holds and zero otherwise. A prediction with multiple action types, or with contradictory directions within the target type, receives no credit for the affected term. Execution failures score zero and remain in the denominator. The metric measures agreement in action type and direction; it does not assess item-identifier accuracy, adjustment magnitude, rationale quality, or downstream business utility.

Given $T$ training batches $B_1,\ldots,B_T$, processed one per evolution cycle, a fixed validation set $V$, and an untouched hold-out set $H$, let
\begin{equation}
 \bar{s}_D(K)=\frac{1}{|D|}\sum_{(x,y^*)\in D} s\big(A_{\phi}(x;K),y^*\big)
 \label{eq:avg}
\end{equation}
denote the observed average score of knowledge $K$ on a set $D$. Let $K_0$ be the initial knowledge, $K_t$ the retained knowledge after cycle $t$, and $\theta$ the minimum validation gain required for acceptance. The goal is to construct $K_T$ through atomic mutations such that each accepted update satisfies
\begin{equation}
 \bar{s}_V(K_t)-\bar{s}_V(K_{t-1})\geq\theta,
 \label{eq:gate}
\end{equation}
while $H$ is never exposed to diagnosis, mutation, or selection. Final effectiveness is the paired gain $\bar{s}_H(K_T)-\bar{s}_H(K_0)$.

\begin{table*}[t]
\centering
\caption{Typed knowledge modules and their intended responsibilities.}
\label{tab:modules}
\setlength{\tabcolsep}{10mm}
\begin{tabular}{ll}
\toprule
Module & Responsibility \\
\midrule
\texttt{role} & Agent identity, objective, and communication stance \\
\texttt{domain} & Advertising products and operating semantics \\
\texttt{experience} & Reusable decision heuristics with stable rule IDs \\
\texttt{constraints} & Evidence, comparison, and safety requirements \\
\texttt{output\_\allowbreak{}protocol} & Analysis termination and structured submission rules \\
\texttt{action\_\allowbreak{}descriptions} & Field-level schema and local-tool descriptions \\
\bottomrule
\end{tabular}
\end{table*}

\begin{table}[t]
\caption{Corpus statistics and domain symbols. Each run uses 800 cases: 480 for training, 160 for validation, and 160 for hold-out evaluation.}
\label{tab:data}
\setlength{\tabcolsep}{5mm}
\begin{tabular}{lccc}
\toprule
Domain & Symbol & Share (\%) \\
\midrule
Home \& Living & $\mathcal{H}$ & 24.8 \\
Automotive & $\mathcal{A}$ & 15.3 \\
Electronics \& Office & $\mathcal{E}$ & 14.7 \\
Fashion \& Apparel & $\mathcal{F}$ & 19.1 \\
Beauty \& Health & $\mathcal{B}$ & 8.2 \\
Long-tail/Others & $\mathcal{L}$ & 17.9 \\
\bottomrule
\end{tabular}
\end{table}

\begin{table*}[t]
\caption{Main results on 160 cases per set (scores $\times100$). Base: $K_0$; KARMA: $K_T$; Gain: absolute points; Rel.: percentage gain over Base. Bold marks the higher score. Avg.\ averages six domains before rounding; symbols follow Table~\ref{tab:data}.}
\label{tab:main}
\begin{tabular}{@{}cccccccccc@{}}
\toprule
 \multirow{2}{*}{LLMs} & \multirow{2}{*}{Domains} & \multicolumn{4}{c}{Validation set $V$} & \multicolumn{4}{c}{Hold-out set $H$} \\
\cmidrule(lr){3-6}\cmidrule(lr){7-10}
 & & Base & KARMA & Gain & Rel.\,(\%) & Base & KARMA & Gain & Rel.\,(\%) \\
\midrule
\multirow{7}{*}{\rotatebox[origin=c]{90}{Qwen3.8-Max}} & $\mathcal{H}$ & 33.13 & \textbf{50.63} & +17.50 & +52.8 & 32.81 & \textbf{48.75} & +15.94 & +48.6 \\
 & $\mathcal{A}$ & 39.06 & \textbf{41.25} & +2.19 & +5.6 & 35.63 & \textbf{36.25} & +0.63 & +1.8 \\
 & $\mathcal{E}$ & 34.38 & \textbf{48.75} & +14.38 & +41.8 & 33.13 & \textbf{43.44} & +10.31 & +31.1 \\
 & $\mathcal{F}$ & 34.69 & \textbf{43.75} & +9.06 & +26.1 & 26.88 & \textbf{38.44} & +11.56 & +43.0 \\
 & $\mathcal{B}$ & 35.31 & \textbf{50.00} & +14.69 & +41.6 & 29.06 & \textbf{47.19} & +18.13 & +62.4 \\
 & $\mathcal{L}$ & 33.75 & \textbf{46.88} & +13.13 & +38.9 & 31.88 & \textbf{47.81} & +15.94 & +50.0 \\
\cmidrule(lr){2-10}
 & Avg. & 35.05 & \textbf{46.88} & +11.82 & +33.7 & 31.56 & \textbf{43.65} & +12.08 & +38.3 \\
\midrule
\multirow{7}{*}{\rotatebox[origin=c]{90}{GLM-5.3}} & $\mathcal{H}$ & 26.88 & \textbf{42.81} & +15.94 & +59.3 & 26.56 & \textbf{40.31} & +13.75 & +51.8 \\
 & $\mathcal{A}$ & 40.00 & \textbf{45.00} & +5.00 & +12.5 & 34.69 & \textbf{36.25} & +1.56 & +4.5 \\
 & $\mathcal{E}$ & 36.25 & \textbf{42.50} & +6.25 & +17.2 & 30.94 & \textbf{39.06} & +8.13 & +26.3 \\
 & $\mathcal{F}$ & 27.81 & \textbf{39.06} & +11.25 & +40.4 & 24.06 & \textbf{36.25} & +12.19 & +50.6 \\
 & $\mathcal{B}$ & 30.00 & \textbf{38.13} & +8.13 & +27.1 & 27.50 & \textbf{39.69} & +12.19 & +44.3 \\
 & $\mathcal{L}$ & 32.19 & \textbf{36.56} & +4.38 & +13.6 & 29.38 & \textbf{35.00} & +5.63 & +19.1 \\
\cmidrule(lr){2-10}
 & Avg. & 32.19 & \textbf{40.68} & +8.49 & +26.4 & 28.85 & \textbf{37.76} & +8.91 & +30.9 \\
\midrule
\multirow{7}{*}{\rotatebox[origin=c]{90}{Kimi-K3}} & $\mathcal{H}$ & 28.13 & \textbf{41.25} & +13.13 & +46.7 & 26.88 & \textbf{36.88} & +10.00 & +37.2 \\
 & $\mathcal{A}$ & 39.38 & \textbf{44.38} & +5.00 & +12.7 & \textbf{40.31} & 37.19 & $-$3.13 & $-$7.8 \\
 & $\mathcal{E}$ & 38.44 & \textbf{41.25} & +2.81 & +7.3 & \textbf{32.81} & 30.94 & $-$1.88 & $-$5.7 \\
 & $\mathcal{F}$ & 31.25 & \textbf{40.63} & +9.38 & +30.0 & 24.69 & \textbf{33.13} & +8.44 & +34.2 \\
 & $\mathcal{B}$ & 30.00 & \textbf{36.88} & +6.88 & +22.9 & 25.94 & \textbf{42.19} & +16.25 & +62.7 \\
 & $\mathcal{L}$ & 28.13 & \textbf{37.50} & +9.38 & +33.3 & 30.31 & \textbf{35.94} & +5.63 & +18.6 \\
\cmidrule(lr){2-10}
 & Avg. & 32.55 & \textbf{40.31} & +7.76 & +23.8 & 30.16 & \textbf{36.04} & +5.89 & +19.5 \\
\bottomrule\end{tabular}
\end{table*}

\begin{table}[t]
\caption{Evidence ablation with Qwen3.8-Max. Bold marks the best score per domain and set.}
\label{tab:ablation}
\small
\setlength{\tabcolsep}{1mm}
\begin{tabular}{@{}clcccccc@{}}
\toprule
\multirow{2}{*}{Domains} & \multirow{2}{*}{Variants} & \multicolumn{3}{c}{Validation set $V$} & \multicolumn{3}{c}{Hold-out set $H$} \\
\cmidrule(lr){3-5}\cmidrule(lr){6-8}
 & & Score & Gain & Rel.\,(\%) & Score & Gain & Rel.\,(\%) \\
\midrule
\multirow{4}{*}{$\mathcal{H}$} & Base & 33.13 & -- & -- & 32.81 & -- & -- \\
 & w/o success & 39.69 & +6.56 & +19.8 & 43.75 & +10.94 & +33.3 \\
 & w/o failure & 38.44 & +5.31 & +16.0 & 35.63 & +2.81 & +8.6 \\
 & KARMA & \textbf{50.63} & +17.50 & +52.8 & \textbf{48.75} & +15.94 & +48.6 \\
\midrule
\multirow{4}{*}{$\mathcal{B}$} & Base & 35.31 & -- & -- & 29.06 & -- & -- \\
 & w/o success & 37.19 & +1.88 & +5.3 & 36.88 & +7.81 & +26.9 \\
 & w/o failure & 39.38 & +4.06 & +11.5 & 39.38 & +10.31 & +35.5 \\
 & KARMA & \textbf{50.00} & +14.69 & +41.6 & \textbf{47.19} & +18.13 & +62.4 \\
\midrule
\multirow{4}{*}{$\mathcal{L}$} & Base & 33.75 & -- & -- & 31.88 & -- & -- \\
 & w/o success & 40.63 & +6.88 & +20.4 & 37.81 & +5.94 & +18.6 \\
 & w/o failure & 41.88 & +8.13 & +24.1 & 38.44 & +6.56 & +20.6 \\
 & KARMA & \textbf{46.88} & +13.13 & +38.9 & \textbf{47.81} & +15.94 & +50.0 \\
\bottomrule
\end{tabular}
\end{table}

\section{Method}
\subsection{System Overview}
Figure~\ref{fig:overview} separates frozen decision execution from knowledge evolution. The inner agent renders current knowledge into its context and response schema, retrieves evidence, and emits a scored decision. The outer loop verifies trajectories, proposes and applies atomic edits, and retains them only after validation. Diagnosis, editing, and acceptance have distinct roles: the Critic proposes changes, the Evolver operates within the knowledge interface, and the deterministic outer loop controls validation and restoration. Accepted knowledge serves the next training batch.

\subsection{Frozen Decision Agent}
The context builder renders knowledge into a prompt $\pi_K$ and response schema $\mathcal{Y}_K$ that remain fixed within each trajectory:
\begin{equation}
 (\pi_K,\mathcal{Y}_K)=\mathrm{Render}(K).
 \label{eq:render}
\end{equation}
A bounded two-stage protocol first collects evidence through white-\\listed read-only business tools, then requires a standalone \texttt{finish\_\\analysis} call before structured output. Each trace records the knowledge fingerprint, tool queries and observations, decision, termination reason, and protocol-error codes $P$. Evidence identifiers $E$ must refer to successful tool responses, and the decision must conform to $\mathcal{Y}_K$. Violations trigger bounded corrective feedback. The trace links the output to the knowledge version and observations used during inference, supporting subsequent inspection of semantic and protocol errors. No advertising action is executed; tool and protocol details appear in Appendix~\ref{app:implementation}.

\subsection{Typed Modular Knowledge}
The knowledge state is a directory of six modules, $K=\{K_m\}_{m\in\mathcal{M}}$, where $\mathcal{M}$ is the set of module identifiers (Table~\ref{tab:modules}); the framework supports controlled edits to all six. Role, domain, and output protocol are typed text modules. Experience and constraints are ordered rule lists,
\begin{equation}
 K_m=\big\langle(\iota_i,\sigma_i,w_i)\big\rangle_{i=1}^{n_m},\qquad m\in\{\mathrm{exp},\mathrm{con}\},
 \label{eq:rules}
\end{equation}
where $\mathrm{exp}$ and $\mathrm{con}$ abbreviate experience and constraints, $n_m$ is the number of rules in module $m$, $\iota_i$ is a stable rule identifier, $\sigma_i$ its source, and $w_i$ its text. Action descriptions parameterize $\mathcal{Y}_K$ and the local finish tool in Equation~\ref{eq:render}.

The registry is the only write boundary. An edit $\mu=(o,m,\iota,v)$ uses operator $o\in\mathcal{O}$ and targets module $m$, optionally rule $\iota$, with value $v$. Operators $\mathcal{O}=\{\textsc{create},\textsc{update},\textsc{delete},\textsc{replace}\}$ support rule edits and text-module replacement. The predicate $\mathrm{Valid}$ checks schema validity across all six modules before writing:
\begin{equation}
 \mathrm{Apply}(K,\mu)=
 \begin{cases}
  \mu(K), & \mathrm{Valid}\big(\mu(K)\big),\\
  K, & \text{otherwise}.
 \end{cases}
 \label{eq:apply}
\end{equation}
Valid changes are written atomically to their target files and journaled with before/after values, and $\mathrm{fp}(K)$ fingerprints the knowledge state.

This representation supports three aspects of controlled adaptation. \emph{Locality}: each mutation names a rule or module field, so its textual scope is explicit. \emph{Auditability}: stable identifiers, provenance, and before/after records connect edits to the diagnostic evidence from their cycle. \emph{Reversibility}: a full-directory snapshot restores the pre-edit knowledge after rejection or error. A cycle permits at most three mutations, which are evaluated together as one candidate patch. Individual edits remain inspectable even when their effects interact during inference.

\subsection{Verification and Full Ingestion}
Without LLM calls, the verifier partitions scored batch trajectories $\mathcal{T}_t$ into full-credit successes $\mathcal{S}_t$, semantic failures $\mathcal{F}_t$, and infrastructure errors $\mathcal{R}_t$. Errors score zero but are reported separately. A semantic failure receives class $c(x)$: \texttt{no\_action} if output is absent, \texttt{wrong\_action\_type} if the type mismatches, and \texttt{wrong\_direction} otherwise.

The candidate-module set $\alpha(x)$ combines two fixed mappings to module sets: $\Gamma$ for failure classes and $\Lambda$ for protocol codes. With $P_x$ denoting the protocol codes recorded for case $x$,
\begin{equation}
 \alpha(x)=\Gamma\big(c(x)\big)\;\cup\bigcup_{p\in P_x}\Lambda(p)\;\subseteq\mathcal{M}.
 \label{eq:attr}
\end{equation}
For example, wrong action types map to domain and experience, where the interpretation of business context and decision heuristics reside. Wrong directions map to experience and constraints; protocol faults map to output protocol or action descriptions. These fixed mappings identify candidate modules for subsequent inspection. They provide a repeatable starting point for the Critic's context-dependent analysis.

Failures and successes are then clustered by
\begin{equation}
 \begin{aligned}
  \kappa_F(x)&=\big(\mathrm{op}(x),c(x),\alpha(x),P_x\big),\\
  \kappa_S(x)&=\big(\mathrm{op}(x),t(y^*_x),d(y^*_x)\big),
 \end{aligned}
 \label{eq:cluster}
\end{equation}
where $\mathrm{op}(x)$ is the record's reference operation, used only after inference, and $y_x^*$ is its reference action. 

The report $\mathcal{D}_t=(\mathcal{F}_t/\kappa_F,\mathcal{S}_t/\kappa_S,\mathcal{R}_t)$ retains sample-level predictions, scores, rationales, evidence identifiers, and compact trace summaries. Here, $\mathcal{F}_t/\kappa_F$ and $\mathcal{S}_t/\kappa_S$ denote partitions into clusters with identical respective keys. Full ingestion covers all batch records. Failures indicate opportunities for repair, while successes expose conditions under which existing knowledge already works. A proposed general rule may match both groups; presenting them together lets the Critic consider that overlap before recommending a change. The evidence-channel ablations test the value of this joint context (Section~\ref{sec:ablation}).

\subsection{Critic and Evolver}
The Critic receives both the complete verifier report and the original trajectories in compact form, and returns a set of edit proposals
\begin{equation}
 \mathcal{P}_t=\mathrm{Critic}(\mathcal{D}_t,\mathcal{T}_t),\qquad
 \mathcal{P}_t=\{(f_k,\delta_k,u_k,m_k,q_k)\}_{k}.
 \label{eq:critic}
\end{equation}
The fields of each proposal denote a failure mode, diagnosis, proposed change, target module, and generality argument, respectively. The Critic compares failures with successes from four perspectives: missing knowledge, conflicting rules, protocol timing, and ambiguous wording. It has no direct knowledge access; the Evolver checks its proposals against current modules before deciding which edits to apply. The outer loop then measures the candidate patch's effect through validation.

The Evolver then turns the report and proposals into a short patch of atomic edits,
\begin{equation}
 \begin{aligned}
  \Delta_t&=\mathrm{Evolver}(K_{t-1},\mathcal{D}_t,\mathcal{P}_t;\mathcal{O}),\\
  \Delta_t&=(\mu_1,\ldots,\mu_k),\qquad 0\leq k\leq M.
 \end{aligned}
 \label{eq:evolver}
\end{equation}
Sequential application through Equation~\ref{eq:apply} produces candidate $\tilde K_t$. The per-cycle mutation budget is $M=3$; the Evolver must inspect current knowledge, avoid sample-specific identifiers, and prefer no edit under uncertainty. The outer loop enforces the mutation budget and tool whitelist. Knowledge inspection and editing are allowed; business tools, shell access, and gate control are not.

\subsection{Validation and Rollback}
Before editing, the outer loop snapshots the knowledge directory. A cycle is \emph{eligible} for gating only if it has semantic failures, a non-empty patch, and a real state change. Its eligibility indicator $e_t\in\{0,1\}$ is
\begin{equation}
e_t=\mathbb{1}\Big[\mathcal{F}_t\neq\emptyset\land |\Delta_t|>0
  {}\land\mathrm{fp}(\tilde{K}_t)\neq\mathrm{fp}(K_{t-1})\Big].
 \label{eq:eligible}
\end{equation}
Here, $|\Delta_t|$ is the number of edits in the patch. Cycles with $\mathcal{F}_t=\emptyset$ skip the Critic and Evolver entirely; skipped or failed evolution cycles have $e_t=0$. For an eligible cycle, the validation gain $g_t=\bar{s}_V(\tilde{K}_t)-\bar{s}_V(K_{t-1})$ instantiates Equation~\ref{eq:gate}, and the knowledge update is
\begin{equation}
 K_t=
 \begin{cases}
  \tilde{K}_t, & e_t=1\land g_t\geq\theta,\\
  K_{t-1}, & \text{otherwise (restore snapshot)},
 \end{cases}
 \label{eq:update}
\end{equation}
with $\theta=0.01$. On rejection or error, restoration returns the six-module knowledge directory to its pre-edit snapshot for subsequent batches.

Validation records for active knowledge are cached: the baseline is evaluated before the first Critic call, an accepted candidate replaces the cache, and rejection preserves it. Candidate inference is performed under the edited knowledge. Caching avoids repeatedly evaluating the unchanged reference state while retaining a fixed comparison for each proposal. Let $N_V$ denote the number of full passes over $V$. Then,
\begin{equation}
 N_V=\mathbb{1}\big[\exists t:\mathcal{F}_t\neq\emptyset\big]+\sum_{t=1}^{T}e_t\leq T+1,
 \label{eq:cost}
\end{equation}
compared with $2T$ passes when naive A/B re-evaluation assesses both states in every cycle. Only after evolution are $K_0$ and $K_T$ evaluated once each on $H$. The gate constrains observed validation regressions and the separate hold-out set assesses whether retained changes generalize.

\section{Experiments}
The experiments address five questions: \textbf{RQ1}, whether knowledge evolution improves held-out decisions across domains and models; \textbf{RQ2}, whether both evidence channels contribute; \textbf{RQ3}, how proposal engines compare under a shared gate; \textbf{RQ4}, whether knowledge transfers across domains; and \textbf{RQ5}, how gated evolution relates to hold-out generalization.

\subsection{Experimental Setup}
\subsubsection{Dataset}
We sample 800 cases per domain from approximately half a million July 2026 operation logs of verified merchants on a global e-commerce platform (Table~\ref{tab:data}). With seed 42, each domain has 480 training cases in six batches of 80, 160 validation cases, and 160 untouched hold-out cases. Logged actions provide behavioral supervision for action selection. The data source and industrial relevance of this study are further discussed in Appendix~\ref{app:industry}.

\subsubsection{Models and Baselines}
We evaluate Qwen3.8-Max, GLM-5.3, and Kimi-K3, using the same family for the inner agent, Critic, and Evolver. Each main setting is run once from shared initial knowledge $K_0$, with $\theta=0.01$ and at most three mutations per cycle. Baselines replace the evolution engine under the same protocol (Section~\ref{sec:engines}); implementation details are in Appendix~\ref{app:implementation}.

\subsubsection{Metrics}
We report average action-agreement scores (Equation~\ref{eq:avg}), absolute gains in points, and relative gains in percent. Scores and absolute gains are scaled by 100; $\theta=0.01$ is one point. Hold-out gains compare $K_T$ with $K_0$ on the same cases. Ablations and engine comparisons use the main-run Base as their common reference; aggregate relative gains use the corresponding mean scores.

\subsection{Main Results}
Table~\ref{tab:main} compares initial and evolved knowledge across 18 domain--model settings. KARMA raises the average hold-out score from 30.19 to 39.15 (+8.96 points, +29.7\%), improving 16 settings. The improvement spans all model families and multiple domains, supporting the usefulness of external knowledge evolution under a fixed executable harness.

All three families gain on average: +12.08 for Qwen3.8-Max, +8.91 for GLM-5.3, and +5.89 for Kimi-K3. Qwen3.8-Max and GLM-5.3 improve in every domain, whereas Kimi-K3 improves in four. Because each family fills the inner-agent, Critic, and Evolver roles, these differences combine decision, diagnosis, and editing ability. Kimi-K3 starts above GLM-5.3 on average but ends below it after evolution, suggesting that stronger initial performance does not necessarily yield greater benefits from knowledge evolution.

Across domains, the largest model-averaged hold-out gains occur in $\mathcal{B}$ (+15.52), $\mathcal{H}$ (+13.23), and $\mathcal{F}$ (+10.73). Benefits are smaller on $\mathcal{E}$ and $\mathcal{L}$, and $\mathcal{A}$ shows little net change. The overall gain therefore coexists with substantial variation in how useful the selected knowledge changes are for each setting.

Kimi-K3 regresses on $\mathcal{A}$ ($-$3.13) and $\mathcal{E}$ ($-$1.88), each after a single early patch. Section~\ref{sec:anatomy} examines the operation-level changes on $\mathcal{A}$. Additional runs with five data seeds for Qwen3.8-Max on $\mathcal{E}$ all yield positive hold-out gains, but their magnitude ranges from +0.94 to +11.25 (Appendix~\ref{app:seeds}). The gains remain positive across these seeds, with substantial variation in magnitude.

\subsection{Evidence Ablations}
\label{sec:ablation}
To assess full ingestion, we remove successful clusters and raw cases (\emph{w/o success}, failure-only reflection) or failed clusters and raw cases (\emph{w/o failure}, success-only consolidation), keeping the loop, gate, and mutation budget fixed. Both variants use Qwen3.8-Max on $\mathcal{H}$, $\mathcal{B}$, and $\mathcal{L}$ with the main run's split and batch order. All gains use the common main-run Base.

\begin{table}[t]
\caption{Evolution engines with Qwen3.8-Max. Tokens: mean total LLM tokens per cycle, including agent inference. Hours: runtime per cycle.}
\label{tab:engines}
\setlength{\tabcolsep}{1mm}
\begin{tabular}{@{}lcccccc@{}}
\toprule
\multirow{2}{*}{Method} & \multicolumn{2}{c}{Validation set $V$} & \multicolumn{2}{c}{Hold-out set $H$} & \multicolumn{2}{c}{Cost} \\
\cmidrule(lr){2-3}\cmidrule(lr){4-5}\cmidrule(lr){6-7}
 & $\mathcal{H}$ & $\mathcal{B}$ & $\mathcal{H}$ & $\mathcal{B}$ & Tokens & Hours \\
\midrule
GEPA & +11.25 & +1.56 & +9.06 & $-$1.88 & 42.15M & 4.58 \\
AHE & +11.56 & +5.94 & +8.75 & +6.56 & 40.74M & 3.40 \\
Self-Harness & +3.75 & +3.13 & $-$2.50 & +2.50 & 40.72M & 3.60 \\
KARMA & \textbf{+17.50} & \textbf{+14.69} & \textbf{+15.94} & \textbf{+18.13} & 41.48M & 4.53 \\
\bottomrule
\end{tabular}
\end{table}

\begin{table}[t]
\caption{Cross-domain hold-out gains with Qwen3.8-Max, relative to each target's main-run Base. Bold marks gains exceeding the target's in-domain gain.}
\label{tab:transfer}
\begin{tabular}{@{}ccccccc@{}}
\toprule
\multirow{2}{*}{Source} & \multicolumn{6}{c}{Target domain} \\
\cmidrule(lr){2-7}
 & $\mathcal{H}$ & $\mathcal{A}$ & $\mathcal{E}$ & $\mathcal{F}$ & $\mathcal{B}$ & $\mathcal{L}$ \\
\midrule
$\mathcal{H}$ & -- & $-$0.63 & +7.81 & \textbf{+17.50} & +12.19 & \textbf{+19.38} \\
$\mathcal{A}$ & $-$5.00 & -- & $-$4.06 & $-$5.94 & $-$1.88 & $-$5.94 \\
$\mathcal{E}$ & +10.63 & \textbf{+9.38} & -- & \textbf{+13.44} & +15.31 & +10.00 \\
$\mathcal{F}$ & +5.94 & $-$3.44 & +8.75 & -- & +10.00 & +13.44 \\
$\mathcal{B}$ & +10.63 & \textbf{+7.19} & +6.88 & \textbf{+11.88} & -- & +15.94 \\
$\mathcal{L}$ & +13.13 & \textbf{+11.88} & +9.06 & \textbf{+19.69} & \textbf{+23.13} & -- \\
\bottomrule
\end{tabular}
\end{table}

\begin{figure}[t]
\centering
\includegraphics[width=\linewidth]{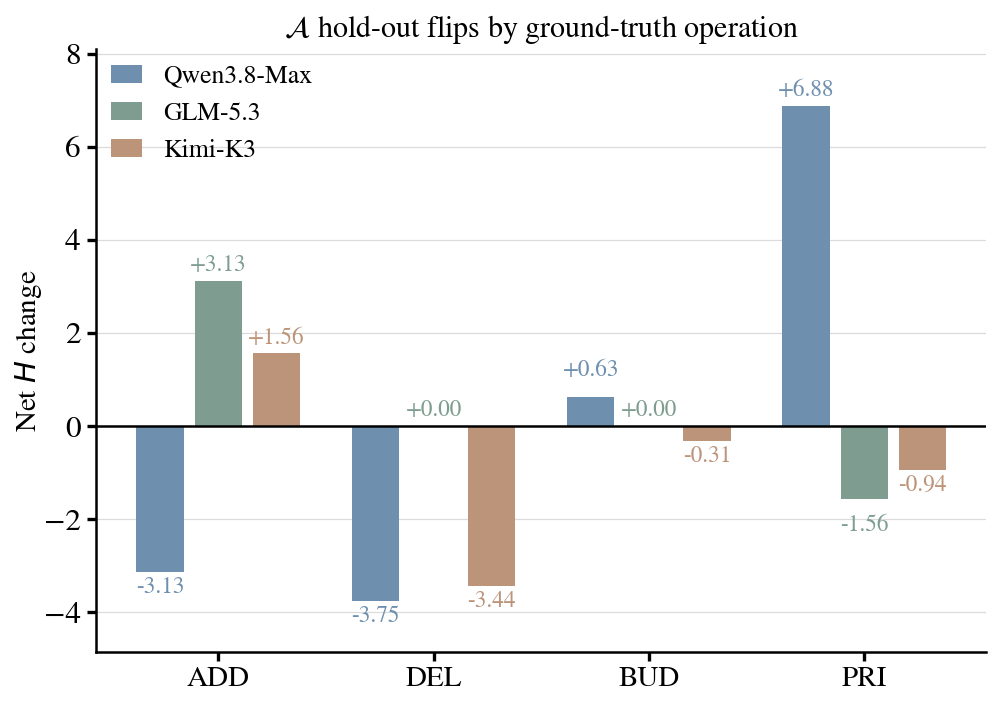}
\caption{Hold-out score changes on $\mathcal{A}$ by reference operation (ADD/DEL: add/delete offer; BUD/PRI: update budget/price). Bars sum to each model's gain in Table~\ref{tab:main}, up to rounding.}
\label{fig:autoanatomy}
\end{figure}

\begin{figure*}[t]
  \centering
  \includegraphics[width=0.9\linewidth]{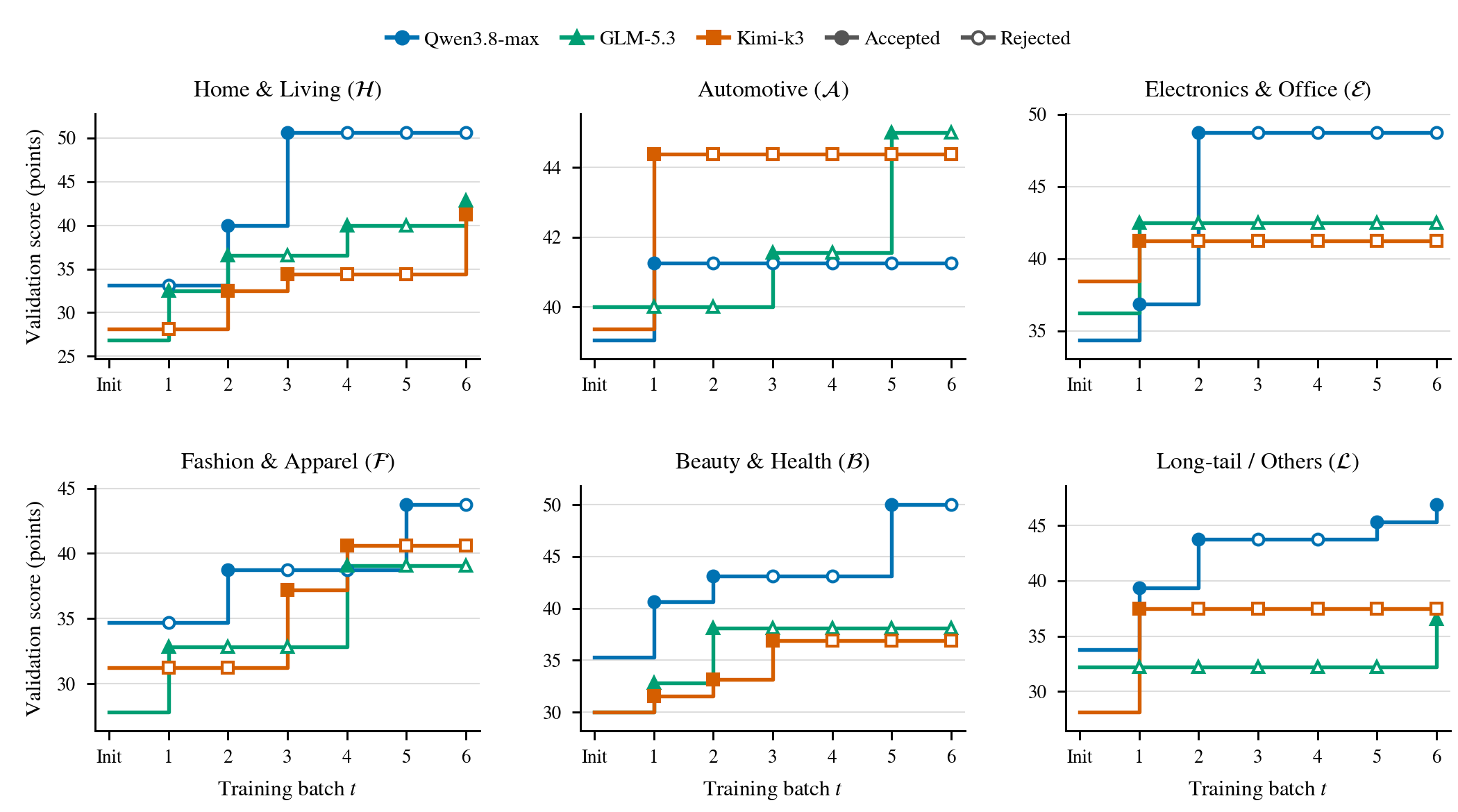}
  \caption{Cached validation scores during evolution (scores $\times100$; Init: $K_0$). Filled markers denote accepted patches; hollow markers denote rejection, which preserves the active knowledge and cached score.}
  \label{fig:evolution}
\end{figure*}

Removing either channel more than halves the average hold-out gain relative to the common Base: +16.67 for KARMA, versus +8.23 without successes and +6.56 without failures (Table~\ref{tab:ablation}). Full ingestion scores highest in every domain on both validation and hold-out sets. The advantage therefore appears in the final evaluation as well as in the scores used for selection.

Neither single channel dominates: failure-only evidence is strong-er on $\mathcal{H}$, success-only evidence on $\mathcal{B}$, and the two are close on $\mathcal{L}$. Success-only consolidation also produces positive hold-out gains in all three domains. Failure evidence directs repair toward incorrect decisions, while successful cases provide examples of behavior to retain or generalize.

\subsection{Evolution Engine Comparison}
\label{sec:engines}
We compare proposal mechanisms while holding the outer loop and validation gate fixed. We replace the Critic and Evolver with GEPA~\cite{gepa}, AHE~\cite{ahe}, or Self-Harness~\cite{selfharness}, re-implemented on the same knowledge directory and mutation interface. All engines use Qwen3.8-Max, the same split and batch order, and $\theta=0.01$. We evaluate on the largest domain $\mathcal{H}$ and smallest $\mathcal{B}$; implementation details are in Appendix~\ref{app:engines}.

KARMA achieves the largest gain in every performance column (Table~\ref{tab:engines}). On $\mathcal{H}$, its hold-out gain is +15.94, compared with +9.06 for GEPA and +8.75 for AHE. On $\mathcal{B}$, it gains +18.13, while the strongest alternative, AHE, gains +6.56. GEPA and Self-Harness show hold-out regressions in one domain each. The engines produce different hold-out outcomes despite sharing the same knowledge interface and acceptance gate.

Mean token use per cycle is similar across engines, while end-to-end runtime varies. KARMA achieves the largest hold-out gains in both domains at a comparable token cost.

\subsection{Cross-Domain Transfer}
\label{sec:transfer}
To examine whether knowledge is tied to its source domain, we freeze the final knowledge evolved by Qwen3.8-Max on each source and evaluate it on the other five domains without further mutation. Each of the 30 source--target entries uses the target's main-run Base hold-out score as its reference (Table~\ref{tab:transfer}). The comparison tests transfer of the knowledge state with the model and target cases held fixed.

The 30 source--target pairs average +8.54 points, about 71\% of the average in-domain gain (+12.08); 23 are positive and nine exceed the target's in-domain gain. Knowledge from $\mathcal{L}$ transfers best to every other domain, averaging +15.38. For example, it gains +23.13 on $\mathcal{B}$, compared with +18.13 from that domain's own evolution. Evolved knowledge is reusable across several product categories, with transfer effectiveness varying by source.

All seven negative entries involve $\mathcal{A}$. Its evolved knowledge lowers the score of every other domain, while $\mathcal{A}$ benefits more from several other sources than from its own evolution. Excluding $\mathcal{A}$ as source or target leaves 20 positive entries averaging +12.73. Automotive is a weak source of transferable knowledge in these runs, even though it benefits from knowledge evolved elsewhere. Section~\ref{sec:anatomy} examines this heterogeneity.

\subsection{Automotive Anomaly}
\label{sec:anatomy}
The small aggregate changes on $\mathcal{A}$ conceal improvements and regressions across operation types (Figure~\ref{fig:autoanatomy}). Qwen3.8-Max gains on price updates while losing on item operations; GLM-5.3 benefits mainly on additions; Kimi-K3's losses concentrate in deletions. Sample-level changes can offset one another, leaving little net gain despite altered decisions.

Mutation records and rationales show new decision priorities alongside existing heuristics. Stable rule identifiers make potential interactions easier to inspect. Transfer results show similar heterogeneity: knowledge evolved on $\mathcal{A}$ harms every other domain, while several other sources improve $\mathcal{A}$ (Table~\ref{tab:transfer}).

\subsection{Gate Dynamics}
\label{sec:gate}
Figure~\ref{fig:evolution} shows stepwise improvements separated by plateaus. Rejected patches leave the active knowledge and cached score unchanged, so these curves track the validation scores of retained knowledge states. Useful updates occur at different stages across models and domains: some runs improve early, while others gain after longer plateaus. Appendix~\ref{app:cases} illustrates accepted and rejected cycles, and Appendix~\ref{app:theta} examines alternative thresholds.

Validation and hold-out improvements are broadly consistent (Table~\ref{tab:main}), but agreement is weaker on $\mathcal{A}$. The gate selects patches using observed validation scores and does not guarantee hold-out improvement. Final evaluation on $H$ provides a separate assessment of generalization.

\section{Limitations and Future Work}
\label{sec:limitations}
Logged actions may encode historical policy biases; agreement does not establish gains in revenue, welfare, or fairness. KARMA is a research system evaluated offline without executing actions. Future work will consider deployment with business guardrails, monitoring, human oversight, and prospective evaluation. Experiments use access controls and anonymized reporting; direct merchant and campaign identifiers are not released.

Deterministic attribution may miss complex interactions, and repeated selection on a fixed validation set can favor changes that do not carry over to hold-out or transfer evaluation. Each main configuration is run once; additional seed, threshold, and comparison studies cover selected settings. Generalization beyond the sampled cases, period, and platform remains open. The framework addresses external knowledge adaptation and cannot repair faulty tools or orchestration code.

\section{Conclusion}
We presented KARMA, a training-free framework for evolving typed, rule-addressable knowledge in advertising agents. Deterministic verification, success/failure reflection, constrained mutation, and validation-gated rollback link retained updates to diagnostic evidence and measured outcomes. Across six domains and three model families, KARMA raises the average hold-out score from 30.19 to 39.15 (+29.7\%). Ablations support using both evidence channels, while cross-domain transfer yields an average gain of 8.54 points. KARMA supports inspectable, reversible knowledge adaptation while keeping model weights and executable harness code fixed.

\bibliographystyle{ACM-Reference-Format}
\bibliography{sample-base}

\appendix

\section{Implementation Details}
\label{app:implementation}
\subsection{Data and Runtime Setup}
Records contain decision context and one historical operation: \texttt{ADD\_OFFER}, \texttt{DELETE\_OFFER}, \texttt{UPDATE\_BUDGET}, or \texttt{UPDATE\_PRICE}, \\mapped to item-add, item-stop, budget-direction, or bid-direction labels. Identifiers support access-controlled queries and are excluded from evolved knowledge. Domain records are shuffled with the run seed before selecting 800 cases; the first 160 form $V$, the next 160 form $H$, and the remaining 480 are shuffled into training batches.

Initial knowledge $K_0$ comes from existing advertising instructions. Thinking is enabled at medium effort; inner inference permits 12 turns and two protocol corrections, and evolution permits eight turns, six Critic proposals, and three mutations. The Critic uses one structured call with at most one formatting retry. Read-only tools retrieve merchant, advertisement, industry, item, and budget-utilization data; failed responses cannot supply evidence identifiers.

\subsection{Diagnostic Evidence}
\label{app:diagnostic}
The verifier report combines a cluster-level overview with the records needed to inspect individual decisions. Each failure cluster contains its frequency, operation, error class, candidate modules, and protocol codes. Its member records retain the prediction, component scores, rationale, evidence identifiers, and execution status. Clusters are ordered by frequency, giving the Critic an overview of recurring errors while retaining access to less frequent cases. Failure and success mode identifiers are local to each batch; rule identifiers persist in the knowledge state.

Table~\ref{tab:attribution} gives the default semantic-error mapping in Equation~\ref{eq:attr}. Protocol codes can add modules to this set. For example, malformed output fields implicate action descriptions, while tool-call and termination faults can implicate output protocol. Successful cases are grouped separately by reference operation, type, and direction. Infrastructure errors have their own report channel and contribute zero to the aggregate score.

\begin{table}[t]
\centering
\caption{Default routing of semantic failures. Protocol codes can add candidate modules to these assignments.}
\label{tab:attribution}
\begin{tabular}{@{}ll@{}}
\toprule
Observed error & Candidate modules \\
\midrule
Missing decision & Output protocol, role \\
Wrong action type & Domain, experience \\
Wrong direction & Experience, constraints \\
\bottomrule
\end{tabular}
\end{table}

Full ingestion preserves the batch membership across these channels. Long interaction traces are represented by compact process summaries containing turn stages, tool names and statuses, termination reasons, and evidence counts. The Critic can therefore compare successful and failed decisions alongside their execution context.

\begin{algorithm}[t]
\caption{KARMA knowledge evolution}
\label{alg:evolution}
\begin{algorithmic}[1]
\Require Initial knowledge $K_0$, batches $B_{1:T}$, sets $V,H$
\Require Threshold $\theta$, mutation budget $M$
\Ensure Final knowledge $K_T$ and hold-out gain
\State $K \gets \mathrm{Copy}(K_0)$; $C_V \gets \bot$
\For{$t=1,\ldots,T$}
  \State $\mathcal{T}_t \gets \mathrm{RunAndScore}(B_t,K)$
  \State $S \gets \mathrm{Snapshot}(K)$
  \State $\mathcal{D}_t \gets \mathrm{Verify}(\mathcal{T}_t)$
  \If{$\mathcal{F}_t=\emptyset$}
    \State \textbf{continue}
  \EndIf
  \If{$C_V=\bot$}
    \State $C_V \gets \mathrm{Evaluate}(V,K)$
  \EndIf
  \State $\mathcal{P}_t \gets \mathrm{Critic}(\mathcal{D}_t,\mathcal{T}_t)$
  \State $(\tilde K,\Delta_t) \gets \mathrm{Evolve}(K,\mathcal{D}_t,\mathcal{P}_t,M)$
  \If{candidate is ineligible}
    \State $K \gets \mathrm{Restore}(S)$; \textbf{continue}
  \EndIf
  \State $\tilde C_V \gets \mathrm{Evaluate}(V,\tilde K)$
  \If{$\mathrm{Mean}(\tilde C_V)-\mathrm{Mean}(C_V)\geq\theta$}
    \State $K \gets \tilde K$; $C_V \gets \tilde C_V$
  \Else
    \State $K \gets \mathrm{Restore}(S)$
  \EndIf
\EndFor
\State $K_T \gets K$
\State Evaluate $K_0$ and $K_T$ once each on $H$
\State \Return $K_T$ and $\bar{s}_H(K_T)-\bar{s}_H(K_0)$
\end{algorithmic}
\end{algorithm}

\subsection{Critic and Evolver Guidance}
\label{app:guidance}
The fixed Critic instructions ask for reusable decision criteria supported by the batch evidence. Each candidate identifies a failure mode and a reflection perspective, explains the diagnosis, proposes a change, names a target module, and states why the proposal should apply beyond the observed cases. The four perspectives focus on missing knowledge, conflicting rules, protocol timing, and ambiguous wording. The Critic is instructed to compare failures with successful cases and to omit candidates when the evidence offers no useful revision.

The Evolver is instructed to inspect the relevant modules before editing and to assess each proposal against the existing knowledge. It may adopt a proposal, adapt it, or reject it with an explanation. Its guidance favors reusable evidence requirements and decision priorities, excludes case-specific identifiers from rule text, and calls for checking knowledge validity after changes. Tool feedback reports invalid arguments or schema violations so that the Evolver can revise its requests. The tool interface enforces the permitted operations and mutation budget; the outer loop determines whether the resulting patch is retained.

\subsection{Engine Implementations}
\label{app:engines}
GEPA reflects on the verifier report and retains a candidate only if it outperforms its parent on eight training cases. AHE requires failure evidence and a predicted effect for each edit, then checks that prediction in the next cycle. Self-Harness proposes a minimal edit for the largest failure cluster and discards it if a second LLM call predicts changes to correctly handled cases. All use the same knowledge interface and outer gate. Token accounting includes engine calls and inference on the 80 training and 160 validation cases of a full cycle; runtime includes final hold-out evaluation.

\subsection{Interfaces and Artifacts}
The interface supports editing all six knowledge modules through \texttt{create\_rule}, \texttt{update\_rule}, \texttt{delete\_rule}, \texttt{update\_text\_module}, and \texttt{update\_action\_description}, with reads, validation, and context preview. The outer loop controls the whitelist, mutation budget, snapshots, and restoration. Run artifacts record split membership, traces, scores, knowledge states, mutation journals, fingerprints, and gate decisions. Identifiers, proprietary tool responses, and credentials are excluded from public reporting.

\section{Evolution Procedure}
\label{app:procedure}
Algorithm~\ref{alg:evolution} uses the fixed split described in Appendix~\ref{app:implementation}. $C_V$ stores validation records and $\mathrm{Mean}(C_V)$ their average score. Here, $\mathcal{F}_t$ is the semantic-failure set identified by $\mathrm{Verify}(\mathcal{T}_t)$, and $\bot$ denotes an uninitialized validation cache. Eligibility follows Equation~\ref{eq:eligible}; an evolution error also makes a candidate ineligible.

\subsection{State Management}
Algorithm~\ref{alg:evolution} maintains three complementary records: a snapshot for restoration, cached validation results for comparison, and a mutation journal for inspection. A fingerprint computed from the normalized module contents identifies changes to the knowledge state. After a successful write or restoration, the knowledge-loading cache is invalidated so that subsequent inference renders the current modules. This loading cache is separate from the validation cache $C_V$, which stores completed evaluation records.

When a candidate is accepted, its knowledge state and validation records become the reference for the next cycle. Rejection restores the snapshot and retains the previous validation reference. The mutation journal is stored outside the knowledge directory and preserves the edited module, target, and before/after contents; restoration is also logged. Thus, a rejected patch remains available for inspection after the active knowledge has been restored.

The cycle record brings together the snapshot, knowledge fingerprints, verifier report, Critic proposals, actual mutations, and gate outcome. Reading these artifacts in sequence connects the evidence that motivated a proposal to the edits applied and the validation decision. Proposal text records the intended change, while the mutation journal records the change actually made.

\section{Robustness Checks}
\label{app:robust}
\subsection{Data Seed}
\label{app:seeds}
We repeat Qwen3.8-Max on $\mathcal{E}$ with seeds 43--46 in addition to 42. Seeds control case selection, partitioning, and batch order, but not LLM sampling; variation therefore combines both sources. All five hold-out gains are positive (+0.94 to +11.25; mean +6.06), with seed 42 toward the upper end (Table~\ref{tab:seeds}).

\begin{table}[t]
\caption{Data-seed variation for Qwen3.8-Max on $\mathcal{E}$ (scores $\times100$). Seed 42$^{\dagger}$ is the main run in Table~\ref{tab:main}. Means are computed before rounding.}
\label{tab:seeds}
\begin{tabular}{@{}lrrrrrr@{}}
\toprule
\multirow{2}{*}{Seed} & \multicolumn{3}{c}{Validation set $V$} & \multicolumn{3}{c}{Hold-out set $H$} \\
\cmidrule(lr){2-4}\cmidrule(lr){5-7}
 & Base & KARMA & Gain & Base & KARMA & Gain \\
\midrule
42$^{\dagger}$ & 34.38 & \textbf{48.75} & +14.38 & 33.13 & \textbf{43.44} & +10.31 \\
43 & 35.31 & \textbf{44.38} & +9.06 & 32.19 & \textbf{43.44} & +11.25 \\
44 & 30.63 & \textbf{35.31} & +4.69 & 31.56 & \textbf{32.50} & +0.94 \\
45 & 30.00 & \textbf{38.44} & +8.44 & 29.38 & \textbf{32.81} & +3.44 \\
46 & 30.63 & \textbf{40.94} & +10.31 & 35.00 & \textbf{39.38} & +4.38 \\
\midrule
Mean & 32.19 & \textbf{41.56} & +9.38 & 32.25 & \textbf{38.31} & +6.06 \\
\bottomrule
\end{tabular}
\end{table}

\subsection{Gate Threshold}
\label{app:theta}
For Qwen3.8-Max on $\mathcal{H}$, $\mathcal{B}$, and $\mathcal{L}$, five of six alternative domain--threshold combinations retain the default gains. Raising $\theta$ reduces the gain on $\mathcal{L}$, but every gain remains positive (Table~\ref{tab:theta}). The shared default $\theta=0.01$ was not selected using hold-out results.

\begin{table}[t]
\caption{Gate-threshold sensitivity with Qwen3.8-Max.}
\label{tab:theta}
\begin{tabular}{@{}lrrrrrr@{}}
\toprule
\multirow{2}{*}{Domain} & \multicolumn{3}{c}{Validation set $V$} & \multicolumn{3}{c}{Hold-out set $H$} \\
\cmidrule(lr){2-4}\cmidrule(lr){5-7}
 & 0.005 & 0.01 & 0.02 & 0.005 & 0.01 & 0.02 \\
\midrule
$\mathcal{H}$ & +17.50 & +17.50 & +17.50 & +15.94 & +15.94 & +15.94 \\
$\mathcal{B}$ & +14.69 & +14.69 & +14.69 & +18.13 & +18.13 & +18.13 \\
$\mathcal{L}$ & +13.13 & +13.13 & +10.00 & +15.94 & +15.94 & +12.50 \\
\bottomrule
\end{tabular}
\end{table}

\section{Case Studies}
\label{app:cases}
\begin{figure*}[t]
  \centering
  \includegraphics[width=\linewidth]{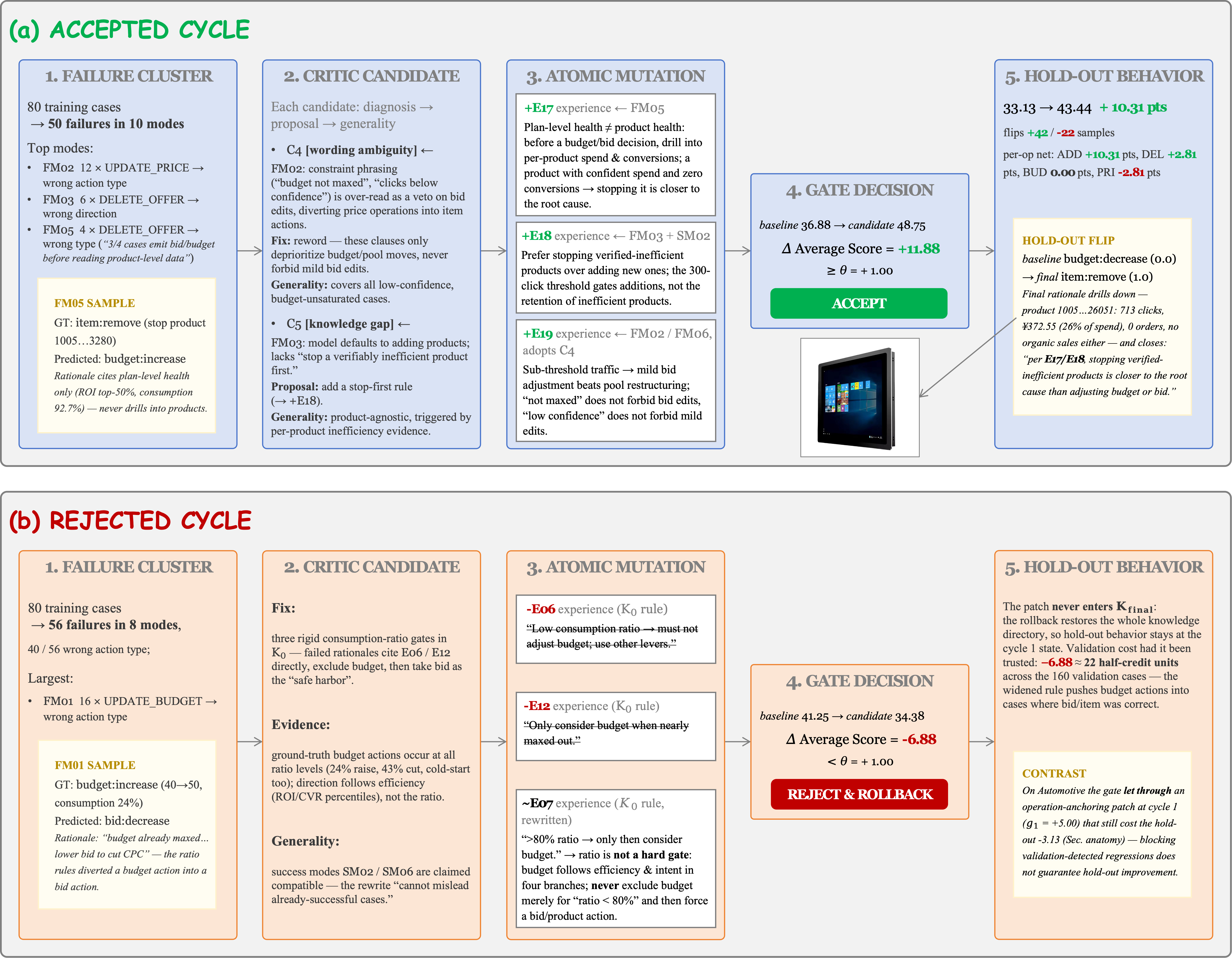}
  \caption{Contrasting evolution cycles on $\mathcal{E}$: (a) Qwen3.8-Max's largest accepted validation gain (+11.88) and (b) Kimi-K3's largest rejected loss ($-$6.88). Each panel follows diagnosis, proposals, mutations, validation against the cached baseline, and hold-out behavior. Final hold-out gains reflect all retained edits. Aggregate scores use the 0--100 scale; sample scores use 0--1; $\theta=1$ point. Rule texts translate the Chinese originals; FM/SM denote batch failure/success clusters. ADD/DEL/BUD/PRI denote the four reference operations.}
  \label{fig:cases}
\end{figure*}

Figure~\ref{fig:cases} shows the largest accepted gain and rejected loss among 108 gated cycles. Both occur at cycle 2 on $\mathcal{E}$.

\paragraph{Accepted cycle.}
For Qwen3.8-Max, the verifier groups 50 failures into ten modes. The Evolver adopts two of six proposals and generalizes them into rules \texttt{E17}--\texttt{E19}, while consulting successes to avoid overbroad changes. Validation improves from 36.88 to 48.75 (+11.88), so the gate retains the patch. Final knowledge, including edits from cycles 1 and 2, improves hold-out performance from 33.13 to 43.44 (+10.31), with both corrected and degraded cases. A corrected rationale explicitly cites \texttt{E17}/\texttt{E18}, illustrating rule-level traceability.

\paragraph{Rejected cycle.}
For Kimi-K3, the verifier finds 56 failures, including 40 wrong action types. Rationales implicate rigid budget-consumption thresholds; the Evolver deletes \texttt{E06} and \texttt{E12} and rewrites \texttt{E07}. Validation falls from 41.25 to 34.38 ($-$6.88), triggering restoration. The rejected patch is never evaluated on $H$, so its hold-out effect is unknown. Conversely, Kimi-K3's accepted patch on $\mathcal{A}$ improves validation but regresses on hold-out (Section~\ref{sec:anatomy}).

The accepted cycle adds decision rules to the retained knowledge, while the rejected cycle restores the previous state after a measured validation loss.

\section{Discussion}
The evidence ablations and engine comparisons examine two design choices: which trajectories inform an update and how proposals are generated. Full ingestion performs best among the evaluated evidence variants, and proposal engines produce different outcomes under a common acceptance rule. Transfer evaluation then measures how the retained knowledge performs in other product domains. These comparisons connect the design of the update process to its observed effectiveness and generalization.

For Web advertising, this separation offers a way to maintain operating knowledge across heterogeneous product categories while preserving the surrounding executable system. Transfer can make a learned rule set useful beyond its source, but the outcomes involving $\mathcal{A}$ show why reuse still requires evaluation in the target setting. Versioned rules and mutation records provide a basis for inspecting both improvements and regressions as those policies evolve.

Evaluation remains offline and measures agreement with logged actions. Experiments use access-controlled data and anonymized reporting. Deployment with monitoring and human oversight is a direction for future work; broader limitations and governance considerations appear in Section~\ref{sec:limitations}.

\section{Industry Relevance}
\label{app:industry}
We submit this work as an Industry Paper because it addresses the practical maintenance of advertising-agent knowledge using real merchant data from AliExpress. Our dataset comprises historical merchant operation logs across six e-commerce domains, covering budget, bid, and advertised-item decisions. These records ground the task and evaluation in actual marketplace operations. KARMA enables operating knowledge to be revised without retraining the model or changing executable code, while traceable edits, validation, and rollback support inspection and controlled updates. The experiments provide offline evidence of improved action agreement and cross-domain knowledge reuse in this industrial setting. Future work will explore deployment with business guardrails, monitoring, and human oversight.

\end{document}